\documentclass{IOS-Book-Article}
\usepackage{multirow}           
\usepackage{graphicx}
\usepackage{amsmath}
\usepackage{pifont}
\usepackage{times}
\usepackage{url}
\normalfont
\usepackage[T1]{fontenc}
\begin{document}
\begin{frontmatter}                           

\title{Efficient Neural Network Approaches for Leather Defect Classification%
}
\runningtitle{An Efficient Neural Network Approach for Leather Defect Classification}

\author[A]{\fnms{Sze-Teng} \snm{Liong}},
\author[B]{\fnms{Y.S.} \snm{Gan}}
,
\author[C]{\fnms{Kun-Hong} \snm{Liu}}
,
\author[D]{\fnms{Tran Quang} \snm{Binh}}
,
\author[D]{\fnms{Cong Tue} \snm{Le}}
,
\author[A]{\fnms{Chien An} \snm{Wu}}
,
\author[A]{\fnms{Cheng-Yan} \snm{Yang}}
,
\author[B]{\fnms{Yen-Chang} \snm{Huang}
\thanks{Corresponding Author: Research Center for Healthcare Industry Innovation, National Taipei University of Nursing and Health Sciences, Taipei, Taiwan R.O.C.; E-mail: 8yenchang@ntunhs.edu.tw.}}

\runningauthor{Liong et al.}
\address[A]{Dept of Electronic Engineering, Feng Chia University, Taichung, Taiwan R.O.C.}
\address[B]{Research Center for Healthcare Industry Innovation, National Taipei University of Nursing and Health Sciences, Taipei, Taiwan R.O.C.}
\address[C]{School of Software, Xiamen University, Xiamen, China}
\address[D]{Faculty of Electrical \& Electronics Engineering, Ton Duc Thang University, Vietnam}

\begin{abstract}
Genuine leather, such as the hides of cows, crocodiles, lizards and goats usually contain natural and artificial defects, like holes, fly bites, tick marks, veining, cuts, wrinkles and others. 
A traditional solution to identify the defects is by manual defect inspection, which involves skilled experts. It is time consuming and may incur a high error rate and results in low productivity. This paper presents a series of automatic image processing processes to perform the classification of leather defects by adopting deep learning approaches.  Particularly, the leather images are first partitioned into small patches, then it undergoes a pre-processing technique, namely the Canny edge detection to enhance defect visualization. Next, artificial neural network (ANN) and convolutional neural network (CNN) are employed to extract the rich image features.  The best classification result achieved is 80.3\%, evaluated on a dataset that consists of$~\sim$ 2000 samples. In addition, the performance metrics such as confusion matrix and Receiver Operating Characteristic (ROC) are reported to demonstrate the efficiency of the method proposed.
\end{abstract}

\end{frontmatter}

\thispagestyle{empty}
\pagestyle{empty}

\section*{Introduction} 
\begin{figure}[t!]
\centering
\includegraphics[width=1\linewidth]{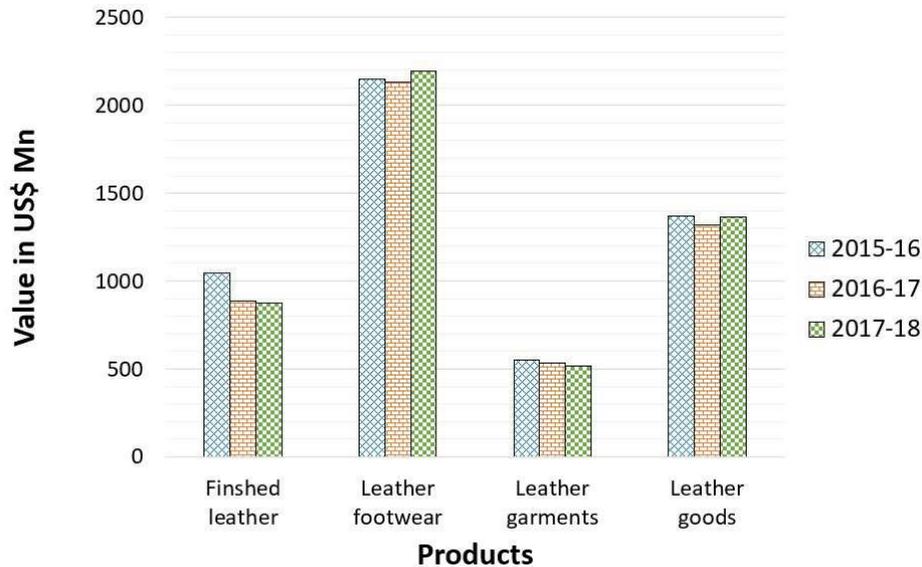}
\caption{India's exports leather products (in US\$ million), for 2015-2018}\label{fig:leather_india}
\end{figure}

According to the statistical studies from Brazil's ministry for external commerce, between January and April 2019, Brazilian exports for the hides and skins reached US \$430 million  ~\cite{brazil}. 
The Brazilian leather industry produced about 15 million ft$^2$ leather in the first four months of 2019, making it the second largest leather production country after China.
On the other hand, India is one of the biggest global exporters of leather especially for footwear and garment products~\cite{india}.
Figure~\ref{fig:leather_india} shows the statistical report for India's exported leather products (in US\$ million), for 2015-2018.
It involves a complex series of treatments to turn hides into leather, which include soaking, pressing, shaving, trimming, dyeing, drying, finishing and selecting.

\begin{figure}[t!]
\centering
\includegraphics[width=1\linewidth]{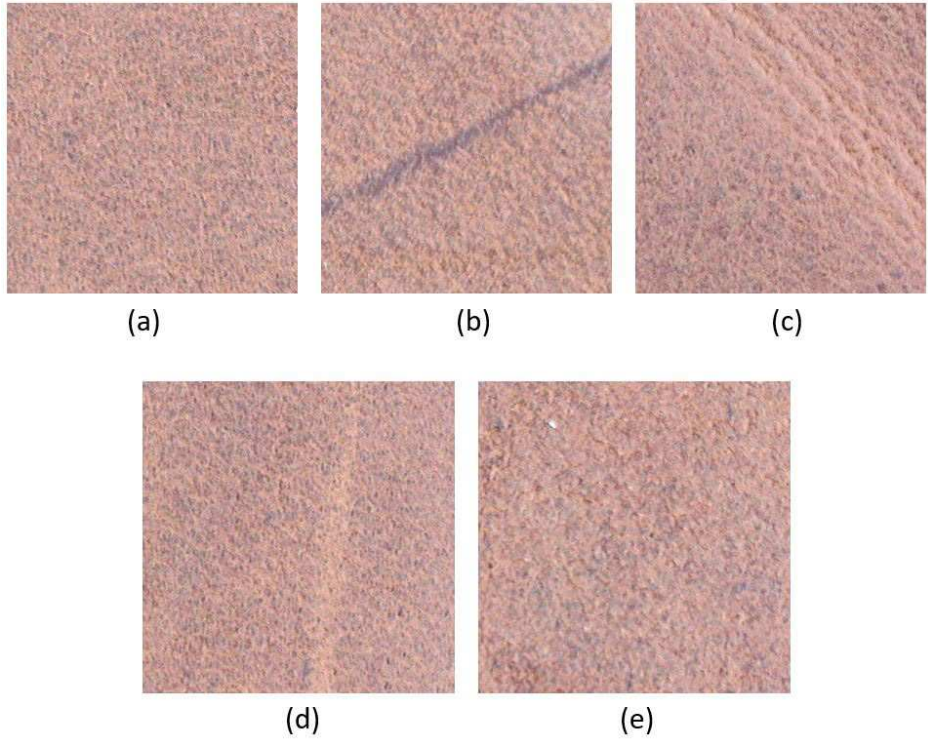}
\caption{Example of leather images: (a) no defect; and with the defects of (b) black line; (c) wrinkle; (d) cuts and (e) stain}\label{fig:defect}
\end{figure}

To produce world-class, quality leather products, it must be ensured that the leather used is defect free.
However, most leather pieces bear the marks of their natural origin, like insect bites, cuts, stains and wrinkles.
The example of the defective images is shown in Figure~\ref{fig:defect}.
These defects should be detected and removed during the filtering process.
To date, the defect detection on leather still relies highly on trained human inspectors. 
It is not reliable and inconsistent as it is highly dependent on the experience of the individual. 
Furthermore, this kind task is repetitive, tedious and physically laborious. 
One could probably spend more time on tasks that require creativity and innovation.
Therefore, an automated quality inspection on leather pieces with digital image processing is essential to assist the defect inspection procedure. 
However, in the literature, there are relatively few researchers from the computer vision field investigating this topic. 

One of the approaches to implement the automation task is by using machine learning. 
Neural network technique allows a computer to behave like a human, particularly in learning and understanding the same way as humans do.
Neural network is gaining a lot of attention in the recent years due to its superior performance.
For example, this technology has been utilized in driverless cars, allowing the cars to automatically recognize a stop sign, or to determine the obstacles on the road. 
Concretely, neural network architectures can achieve state-of-the-art accuracy in many classification tasks and sometimes even exceed human-level performance, such as in speech recognition~\cite{xiong2018microsoft} and object recognition~\cite{he2016deep}.
The neural network model is a set of algorithms and is usually trained by a large set of labeled data.
It requires high-performance GPUs with parallel architecture to increase the computational speed.

\begin{figure*}[t!]
\centering
\includegraphics[width=1\linewidth]{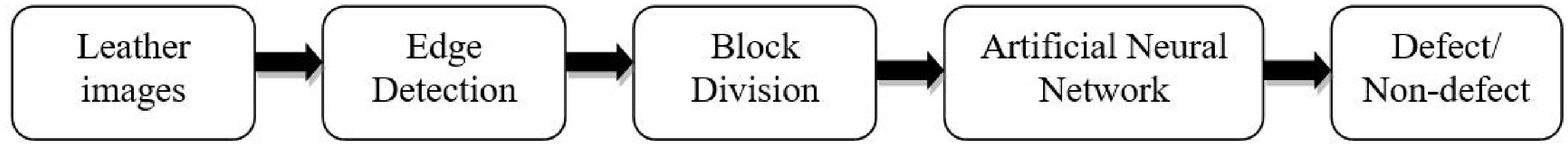}
\caption{Flowchart of the leather defect classification using neural network}\label{fig:flow}
\end{figure*}

This paper attempts to propose an image processing technique for leather classification by employing the neural network method.
The leather images are pre-processed using edge detection and block partition, before performing the feature extraction and classification with neural network.
The overview of the proposed method is shown in Figure~\ref{fig:flow}.
The rest of this paper is organized as follows: Section~\ref{sec:literature} discusses a brief review of related literature, followed by Section~\ref{sec:method} which describes the proposed method in detail.
Next, Section~\ref{sec:results} summarizes the experimental results while the conclusion
is drawn in Section~\ref{sec:conclusion}.

\section{Literature Review} \label{sec:literature}
One of the recent research about leather classification is presented by Bong et al.~\cite{bong2018vision}. 
They employed several image processing algorithms to extract the image features and identify the defect's position on the leather surface. 
The extracted features (i.e., color moments, color correlograms, Zernike moments, and texture) are evaluated on an SVM classifier. 
Total number of image samples collected are 2500, where 2000 are used as training data and 500 samples are used as the testing data.
The testing accuracy in distinguishing the three types of defects (scars, scratches, pinholes) as well as also no defect is 98.8\%.
However, such method requires camera environment to be setup and static for consistent leather images, and might be time consuming for finding best parameters for model training.

Jawahar et al.~\cite{jawahar2014leather} proposed a wavelet transform to classify the leather images.
They adopt the Wavelet Statistical Features (WSF) and Wavelet Co-occurrence Features (WCF)~\cite{jobanputra2004texture} as feature descriptors.
There are a total of 700 leather images involved, including 500 defective and 200 non-defective samples.
The dataset is partitioned into 2 parts, where 70\% is the train set and 30\% is the test set.
A binary SVM with Gaussian kernel is exploited to differentiate the defective and non defective leather sample. 
The classification accuracy of WSF, WCF and WSF+WCF are 95.76\%, 96.12\% and 98.56\%, respectively.
However, the description of the defect types is unknown as an obvious visualization of the defect is easier for analysis and classification.

On the other hand, Pistori et al.~\cite{pistori2018defect} presented Gray-scale Coocurrence Matrix (GLCM)~\cite{jobanputra2004texture} to extract the features of the images. 
The dataset is elicited from 258 different pieces of raw hide and wet blue leather, and they contain 17 different defect types. 
For the experiment, four types of defects are chosen, namely, tick marks, brand marks made from hot iron, cuts and scabies.
Ridge estimators and logistic regression are adopted to learn the normalized Gaussian radial basis functions.
They are then clustered by SVM, Radial Basis Functions networks (RBF) and Nearest Neighbours (KNN) as classifiers.
Among them, SVM achieved the best results: beyond 94\% by using 10$\times$10 window image size and 100\% when 40$\times$40 window size is considered.

Another leather detection work is carried out by Pereira et al.~\cite{pereiraclassification}.
A Pixel Intensity Analyzer (PIA) is employed as the feature descriptor with Extreme Learning Machines (ELM) as the classifier.
It describes the entire process going from image acquisition to image pre-processing, features extraction and finally machine learning classification. 
However the paper did not describe the machine setup it used to run the experiment. 
The performance comparison might be different on different machines.

Winiarti et al.~\cite{winiarti2018pre} aims to realise an automatic leather grading system.
At this stage, it classifies the type of leather on tanning leather images. 
It uses the first seven layers of AlexNet to extract features from the images and then classifies the images using linear SVM. 
The proposed method performs better than using a hand-crafted feature extractor (colour moments + GLCM) combining with SVM classifier in term of its average accuracy, specificity, sensitivity, precision, and performance time.

Our previous work~\cite{liong2019automatic} discusses the elicitation of the dataset in detail and locates the fly bite defect with a segmentation accuracy of 91\%. 
The experiment is examined on a relatively small dataset that only contains 584 images.
In~\cite{liong2019integrated}, ANN is employed to extract features and classify the image. 
The input images are resized to 40$\times$40 and edge detection methods such as Canny, Prewitt, Sobel, Roberts, LoG and ApproxCanny are performed.
The classification result is 82.5\% when the number of hidden neurons is set to 50. 
Note that there is a single hidden layer in their implementation.

\section{Method Proposed}\label{sec:method}
For the feature extractors, there exists handcrafted methods (i.e., statistical features) and neural network approaches (i.e., AlexNet architecture).
We propose a method using both the Convolutional Neural Network (CNN) and ANN as the feature extractors and classifiers to differentiate the defective/ non-defective leather images.
The details are described in the subsections: Section~\ref{sec:ANN} to explain about ANN approach, and Section~\ref{sec:CNN} to elaborate CNN method.

\subsection{Dataset}
\begin{figure}[t!]
\centering
\includegraphics[width=1\linewidth]{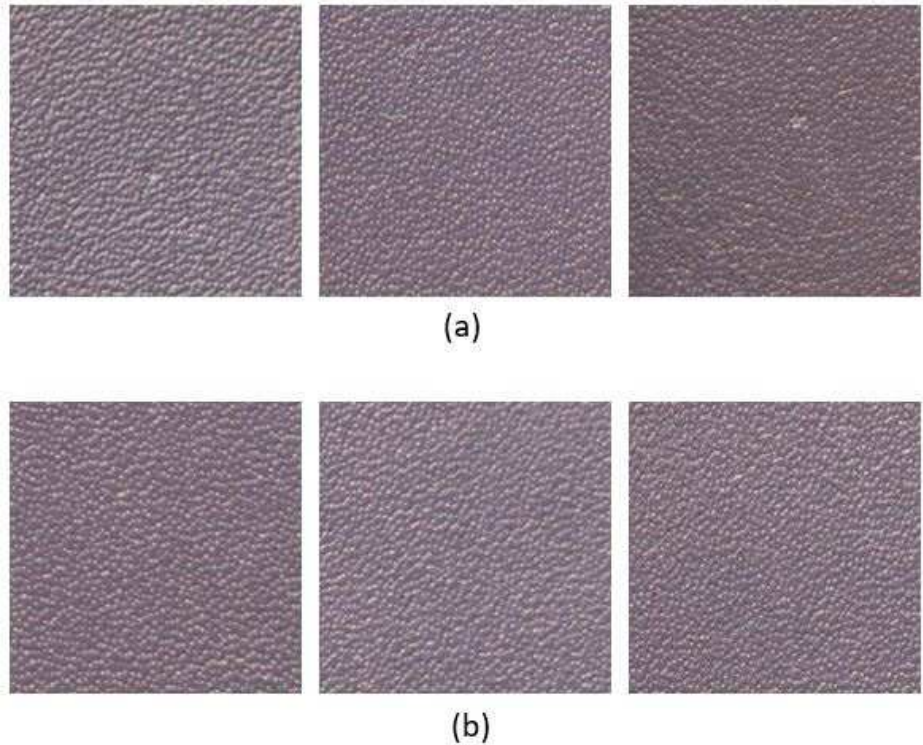}
\caption{The sample of leather images, that has (a) defect and (b) no defect}\label{fig:sample}
\end{figure}

A new dataset is created by collecting the leather images using a six-axis articulated robot DRV70L Delta.
To avoid the flicker caused by the fluorescent lights, a professional lighting source is used, when capturing the images using a DSLR camera.
The amount of the data collected is 1897 images, where each of them are 400$\times$400 pixels.
There are 370 images containing the fly bite defect and the rest are non-defective.
The sample images are shown in Figure~\ref{fig:sample}.
More information about the elicitation of the dataset can be found in~\cite{liong2019automatic, liong2019integrated}.

\subsection{Artificial Neural Network}\label{sec:ANN}
A few image pre-processing techniques are applied on the leather images prior to passing them into the ANN for feature extraction.

\subsubsection{Data Pre-processing}

\begin{figure}[t!]
\centering
\includegraphics[width=1\linewidth]{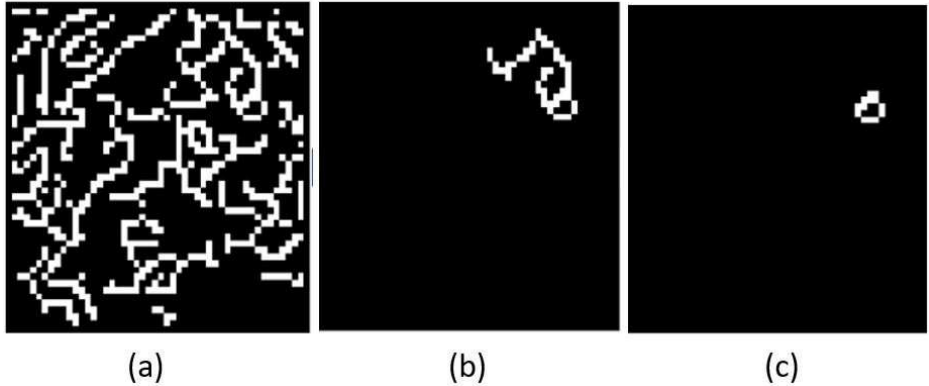}
\caption{After processing the leather image of Canny edge detection with the threshold range of (a) [0, 1], (b) [0.2, 0.9] and (c) [0.5, 0.9] }\label{fig:canny}
\end{figure}

\begin{figure}[t!]
\centering
\includegraphics[width=0.8\linewidth]{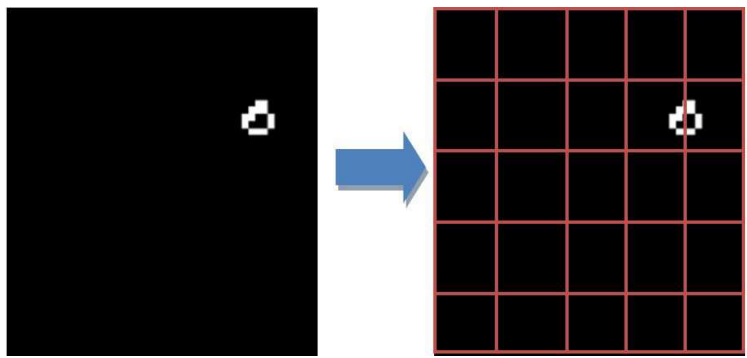}
\caption{The image is partitioned into 5$\times$5 blocks}\label{fig:block}
\end{figure}

All the images are put through four pre-processing steps, namely, RGB to grayscale, re-sizing, edge detection and block partition.
These steps are to enhance the visibility of defective regions in the images. 
Succinctly, the original images have a resolution of 400 $\times$ 400 $\times$ 3.
They are first converted to grayscale, therefore becoming 400 $\times$ 400 $\times$ 1.
Next, the images are re-sized to 50 $\times$ 50 pixels.
There are several options for edge detection methods, such as Sobel, Prewitt, Roberts and Canny.
The edge detector that can display the leather defect most clearly is the Canny operator. 
By adjusting the threshold values of the operator, different effects are obtained, as illustrated in Figure~\ref{fig:canny}.
Finally, each image is divided into 5$\times$5 blocks, as shown in Figure~\ref{fig:block}.
Since the pixel intensity of each image is either 0 or 255, the frequency of occurrence of pixel values are calculated.
Thus, each block of the image will form 2 values and each image has 50 feature vectors.

\subsubsection{Feature Extraction using ANN}
ANN consists of an interconnected group of nodes that link the three basic layers of neurons.
Classically, the layers include input, hidden and output. 
Each ANN has one input layer, one output layer and may include more than one hidden layer.
The neurons derive a unique pattern from the image and make decisions based on the extracted features.
The number of neurons in the input layer in this study is fixed to 50 and there is one neuron in the output layer (i.e., either 0/1, which is defect/ no defect).

\subsubsection{Experiment Configuration for ANN}

\begin{figure}[t!]
\centering
\includegraphics[width=0.7\linewidth]{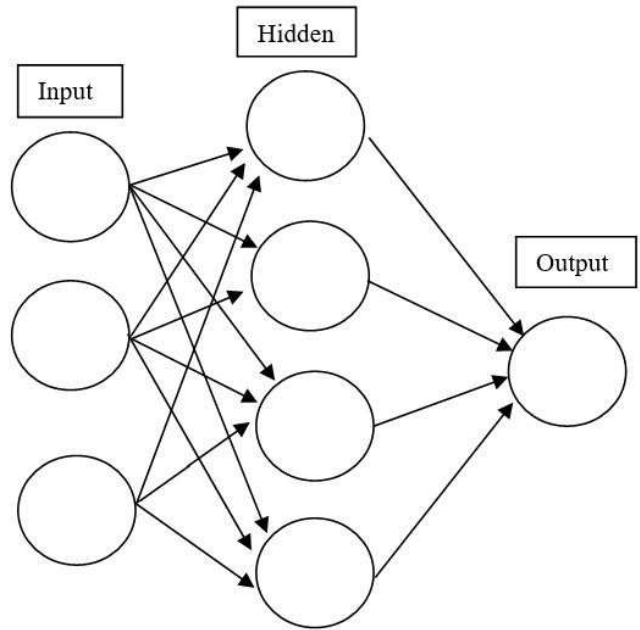}
\caption{Example of Artificial Neural Network with one input, hidden and output layer, respectively.}\label{fig:ann}
\end{figure}

The data is split to three sets: training, validation and testing.
This is to ensure that there is no overlapping of the images.
The training samples are fed into the architecture to adjust the parameters (i.e, weights and biases) to best describe the features of the input data.
Validation samples give clues regarding the network generalization to prevent the architecture from overfitting or underfitting.
The testing samples are to evaluate the classification performance of the unseen input data to examine the robustness of the trained architecture.
Concisely, among the 1897 images,  60\% of them (1138 samples) are allocated for training, 5\% (95 samples) is the validation set and 35\% (664) is the testing set. 
The example of the ANN is shown in Figure~\ref{fig:ann}.

\subsection{Convolutional Neural Network}\label{sec:CNN}
As CNN is capable to extract low-level features (i.e., lines, edges, curves), mid-level features (i.e., circles, squares) and high-level features (shapes and objects).
The images are pre-processed by simply performing a resize operation.

\subsubsection{Data Pre-processing}
To reduce the computational speed while maintaining the image quality, we attempt to decrease the spatial resolution of the original image.
Concretely, the images are resize to 50$\times$50$\times$3, 100$\times$100$\times$3, 150$\times$150$\times$3 and 200$\times$200$\times$3, from the original size of 400$\times$400$\times$3.
	
Due to the distribution of the image dataset is imbalance, we build three sub-datasets by randomly selecting the images from the image collected.
Concisely, there will be 1:1, 1:2 and 1:3 ratios for the defective:non-defective images, respectively.
We further remove the images that are irrelevant to the fly bites defect, such as the images with wrinkles, stains and some blur images.
As a result, the remaining amount of defective images left is 233.
Next, the images are categorized to ``bright" and ``dark" groups. 
They are distinguished by the sum of the pixel intensity values in the image.
For instance, if more than 70\% of the pixels in an image are greater than the intensity value of 125, it is defined as ``bright" image; otherwise it is the ``dark" image.
Consequently, there are 92 and 141 defective images that are bright and dark, respectively.
			
\subsubsection{Feature Extraction using CNN}
A pre-trained neural network (i.e., AlexNet) is utilized with slight modification.
The details structure of the modified AlexNet architecture is tabulated in Table~\ref{table:alex}, for the input image of 150$\times$150$\times$3.
Basically, the architecture comprised of five types of operation: convolution, ReLU, pooling, fully connected and dropout:

\begin{enumerate}
    \item Convolution: The image performs a dot product between a kernel/ weight and the local regions of the image.
    This step can achieve blurring, sharpening, edge detection, noise reduction effect.
    
    \item ReLU: An element-wise activation function is applied as thresholding a technique, such as $max(0,x)$.
    This is to eliminate the neurons that are playing vital role in in discriminating the input and is essentially meaningless.
    
    \item Pooling: To downsample the image along the spatial dimensions (i.e., width and height).
    This allows dimension reduction and enables the computation process to be less intensive.
    
    \item Fully connected: All the previous layer and next layer of neurons are linked.
    It acts like a classifier based on the features from previous layer.
    
    \item Dropout: The neurons are randomly dropped out during the training phase.
    This can avoid the overfitting phenomena and enhance the the generalization of the neural network trained.
    
\end{enumerate}

Specifically, the parameters of the input and output layers are changed, while the other layers remain the same.

\subsubsection{Experiment Configuration for CNN}
Since the dataset to be evaluated in this section is lesser compared to ANN. 
A conventional machine learning approach is employed, viz., k-fold cross-validation (CV).
The general procedure of implementing the k-fold CV is: 
(1) The dataset is shuffled randomly;
(2) The dataset is then split into k subsets;
(3) A subset is selected as the test set, whereas the rest are the training sets;
(4) The model is trained on the training set and evaluate om the test set;
(5) The evaluation score is recorded and the model trained is discarded;
(6) Steps 3 to 5 is repeated;
(7) All the k sets of evaluation scores are summarized to form the final classification accuracy.
Particularly, we fix the value of k to 10 in the experiment.

\begin{table*}[tb] 
\begin{center}
\caption{Modified AlexNet architecture for leather defect classification}
\label{table:alex}
\begin{tabular}{lccccccc}
\noalign{\smallskip}
\hline
\noalign{\smallskip}
Layer
& Filter/ pool size
& \# filter
& Stride
& Padding
& Channel/ element
& \%
& Output size \\
\hline

\noalign{\smallskip}
Input image
& -
& -
& -
& -
& -
& -
& 150 $\times$ 150 $\times$ 3 \\

\noalign{\smallskip}
Convolution 1
& 11 $\times$ 11$\times$ 3
& 96
& [4, 4]
& [0, 0, 0, 0]
& -
& -
& 35 $\times$ 35 $\times$ 96 \\

\noalign{\smallskip}
ReLU 1
& -
& -
& -
& -
& -
& -
& 35 $\times$ 35 $\times$ 96 \\

\noalign{\smallskip}
Normalization 1
& -
& -
& -
& -
& 5
& -
& 35 $\times$ 35 $\times$ 96 \\

\noalign{\smallskip}
Pooling 1
& 3 $\times$ 3
& -
& [2, 2]
& [0, 0, 0, 0]
& -
& -
& 17 $\times$ 17 $\times$ 96 \\

\noalign{\smallskip}
Convolution 2
& 5 $\times$ 5$\times$ 48
& 256
& [1, 1]
& [2, 2, 2, 2]
& -
& -
& 17 $\times$ 17 $\times$ 256 \\

\noalign{\smallskip}
ReLU 2
& -
& -
& -
& -
& -
& -
& 17 $\times$ 17 $\times$ 256 \\

\noalign{\smallskip}
Normalization 2
& -
& -
& -
& -
& 5
& -
& 17 $\times$ 17 $\times$ 256 \\

\noalign{\smallskip}
Pooling 2
& 3 $\times$ 3
& -
& [2, 2]
& [0, 0, 0, 0]
& -
& -
& 8 $\times$ 8 $\times$ 256 \\

\noalign{\smallskip}
Convolution 3
& 3 $\times$ 3 $\times$ 256
& 384
& [1, 1]
& [1, 1, 1, 1]
& -
& -
& 8 $\times$ 8 $\times$ 384 \\

\noalign{\smallskip}
ReLU 3
& -
& -
& -
& -
& -
& -
& 8 $\times$ 8 $\times$ 384 \\

\noalign{\smallskip}
Convolution 4
& 3 $\times$ 3 $\times$ 192
& 384
& [1, 1]
& [1, 1, 1, 1]
& -
& -
& 8 $\times$ 8 $\times$ 384 \\

\noalign{\smallskip}
ReLU 4
& -
& -
& -
& -
& -
& -
& 8 $\times$ 8 $\times$ 384 \\

\noalign{\smallskip}
Convolution 5
& 3 $\times$ 3 $\times$ 192
& 256
& [1, 1]
& [1, 1, 1, 1]
& -
& -
& 8 $\times$ 8 $\times$ 256 \\

\noalign{\smallskip}
ReLU 5
& -
& -
& -
& -
& -
& -
& 8 $\times$ 8 $\times$ 256 \\

\noalign{\smallskip}
Pooling 5
& 3 $\times$ 3
& -
& [2, 2]
& [2, 2, 2, 2]
& -
& -
& 5 $\times$ 5 $\times$ 256 \\

\noalign{\smallskip}
Fully Connected 6
& 4096 $\times$ 6400
& -
& -
& -
& -
& -
& 4096 $\times$ 1 \\

\noalign{\smallskip}
ReLU 6
& -
& -
& -
& -
& -
& -
& 4096 $\times$ 1 \\

\noalign{\smallskip}
Dropout 6
& -
& -
& -
& -
& -
& 50
& 4096 $\times$ 1 \\

\noalign{\smallskip}
Fully Connected 7
& 4096 $\times$ 4096
& -
& -
& -
& -
& -
& 4096 $\times$ 1 \\

\noalign{\smallskip}
ReLU 7
& -
& -
& -
& -
& -
& -
& 4096 $\times$ 1 \\

\noalign{\smallskip}
Dropout 7
& -
& -
& -
& -
& -
& 50
& 4096 $\times$ 1 \\

\noalign{\smallskip}
Fully Connected 8
& 2 $\times$ 4096
& -
& -
& -
& -
& -
& 2 $\times$ 1 \\

\noalign{\smallskip}
Output
& -
& -
& -
& -
& -
& -
& 2 $\times$ 1 \\
\hline

\end{tabular}
\end{center}
\end{table*}

\section{Results and Discussion}
\label{sec:results}
Since there are two methods evaluated on the dataset: ANN and CNN.
We report and discuss both the classification performances in Section~\ref{sec:res_ANN} and Section~\ref{sec:res_CNN}.

\subsection{Results for ANN}
\label{sec:res_ANN}

Several numbers of hidden neurons are tested (i.e., 10, 20, 50 and 100).
The experimental results for the effectiveness of the pre-processing steps are presented in Table~\ref{table:result}.
It can be observed that the performance after performing the edge detection and block division is higher. 
The edge detection employed here is Canny and the threshold is set to [0.5, 0.9]; this is because the defect in the image can be more obvious by using this range.
The best classification result is 80.3\%, where the number of neurons in the hidden layer is 50.
Within the range tested, good accuracy is consistently achieved when the number of hidden neurons is set to 50.

\begin{figure}[t!]
\centering
\includegraphics[width=1\linewidth]{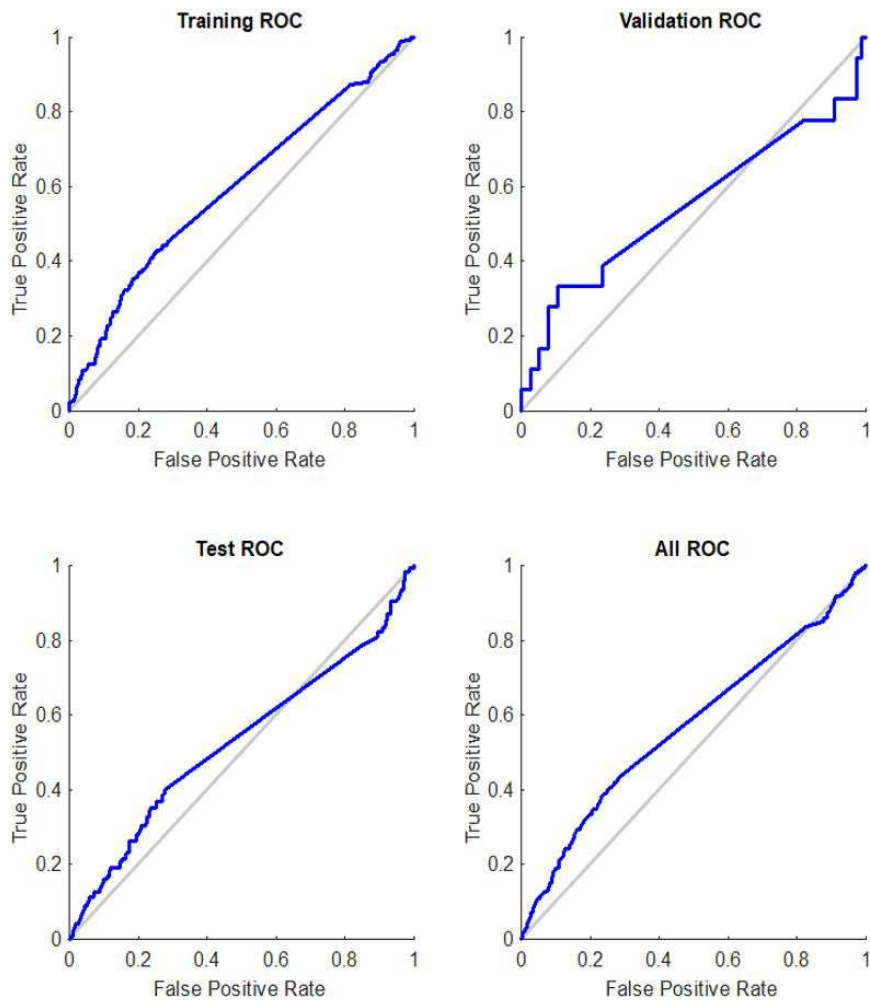}
\caption{Receiver Operating Characteristic for the performance in training, validation and testing sets}\label{fig:roc}
\end{figure}

The confusion matrix of the highest result is shown in Table~\ref{table:confusion}.
It is noticed that there are imbalance issues in the test set, as the total number of defective samples is 9, where about one-third of them can be classified correctly. 
In contrast, there are 655 samples of non-defective images and most of them (more than 80\%) are distinguishable by the trained architecture.
In addition, its ROC is reported in Figure~\ref{fig:roc}, which further indicates the severe imbalanced class distribution problem.

\begin{table}[!tb] 
    \caption{Classification results when varying neurons of hidden layer}
      \label{table:result}
      \centering
        \begin{tabular}{ccccc}
        \noalign{\smallskip}
        \cline{2-5} 
        \noalign{\smallskip}
        & \multicolumn{4}{c}{Number of neuron}
        \\
        \noalign{\smallskip}
        \cline{1-5} 
        \noalign{\smallskip}
        Proposed Method &	10  & 20 & 50 & 100	\\
        \noalign{\smallskip}
        \hline
        \noalign{\smallskip}
        W/o both edge detection \& block division 
        & 78.0	& 79.2	& 80.2 & 79.5	\\
        With edge detection \& w/o block division  
        &79.2 & 78.5 &  80.0 & 78.8 	\\
        With both edge detection \& block division    
        &78.6 & 78.8 & \bf 80.3 &  79.3	\\
        \hline
        \end{tabular}
\end{table}

\begin{table}[!tb] 
    \caption{Confusion matrix by adopting the ANN approach for the test set}
      \label{table:confusion}
      \centering
        \begin{tabular}{lcc}
        \noalign{\smallskip}
        \cline{1-3} 
        \noalign{\smallskip}
         &	Non-defective   & Defective 	\\
        \noalign{\smallskip}
        \hline
        \noalign{\smallskip}
         Non-defective
        & 530	& 125 	\\
         Defective 
        & 6 & 3   	\\
        \hline
        \end{tabular}
\end{table}

\subsection{Results for CNN}\label{sec:res_CNN}
The images are carefully selected from the dataset collected before performing the evaluation using the modified AlexNet.
The distribution of the data subsets is tabulated in Table~\ref{table:datadistribution}.
For example, there will be a total of 932 ``bright+dark" images involved during the 10-fold CV for 1:3/ defective:non-defective case.
In brief, 233 of them are defective images and 699 images have no defect.
Among them, 10\% (i.e., 47) of the images are treated as the test set and 90\% (i.e., 419) are the training set.
On the other hand, the minimum number of images for one of the cases is 184, which is when considering only the dark images for 1:1 case.

The classification results are reported in Tables~\ref{table:cnn_11},~\ref{table:cnn_12} and ~\ref{table:cnn_13}, for the 1:1, 1:2 and 1:3 data subsets, respectively.
The highest result obtained is 76.2\% in the bright images when epoch = 180 and resolution = 150$\times$150, in the 1:3 case.
The lowest result attained is 50\%, which is as good as a chance, as there are only two target classes.
It is in the 1:2 case, when epoch = 100 and resolution = 50$\times$50.

It is observed that 50$\times$50 are always underperformed in most of the cases, compared to other input resolutions.
There is a trend indicates that the classification accuracy is higher when the dataset is larger.
For instance, the results in Table~\ref{table:cnn_11} (i.e., 1:1) is lower compared to Table~\ref{table:cnn_13} (i.e., 1:3).
The epoch number set here is in the range of [100, 200], which is considered relatively small compared to general classification task.
It implies that, with a slight fine-tuning on the architecture parameters (i.e., weights and biases) are sufficient to encode important features of the leather images.

The confusion matrices for the highest results achieved in the 1:3 case for the ``bright+dark", ``bright" and ``dark" cases are reported in Tables~\ref{table:confusion_13_brightdark},~\ref{table:confusion_13_bright} and~\ref{table:confusion_13_dark}.
It can be seen that the although the classification accuracy of Table~\ref{table:confusion_13_dark} reaches 74\%, the are $\sim$75\% (i.e., 69 out of 92 images) of the defective images are being predicted wrongly, whereas  $\sim$90\% (i.e., 
250 out of 276 images) non-defective images are correctly classified.

\begin{table}[!tb] 
    \caption{Data distribution for evaluation using modified AlexNet}
      \label{table:datadistribution}
      \centering
        \begin{tabular}{lccc}
        \noalign{\smallskip}
        \cline{2-4} 
        \noalign{\smallskip}
        
         &	 \multicolumn{3}{c}{Defective: Non-defective} 	\\
        \noalign{\smallskip}
        \hline
        \noalign{\smallskip}
        & 1:1 
        & 1:2
        & 1:3 \\ \hline
        
        Bright + Dark
        & 233 : 233
        & 233 : 466 
        & 233 : 699 \\
        
        Bright 
        & 141 : 141
        & 141 : 282 
        & 141 : 423 \\
        
        Dark
        & 92 : 92
        & 92 : 184 
        & 92 : 276 \\
        
        \hline
        \end{tabular}
\end{table}

\begin{table}[!tb] 
    \caption{Classification results for 1:1 data subset when varying the epoch value using modified AlexNet}
      \label{table:cnn_11}
      \centering
        \begin{tabular}{cccccccc}
        \noalign{\smallskip}
        \cline{3-8} 
        \noalign{\smallskip}
        
        & & \multicolumn{6}{c}{Epoch}\\
        
        \noalign{\smallskip}
        \cline{2-8} 
        \noalign{\smallskip}
         & Resolution 
         & 100 
         & 120
         & 140
         & 160
         & 180
         & 200	\\
        \noalign{\smallskip}
        \hline
        \noalign{\smallskip}
        
        \multirow{4}{*}{Bright + Dark} 
        & 50$\times$50
        & 58.1
        & 58.5
        & 59.2
        & 58.3
        & 58.7
        & 58.3\\
        
        & 100$\times$100
        & 62.4
        & 58.5
        & 61.3
        & 60.5
        & 64.5
        & 61.8 \\
        
        & 150$\times$150
        & 62.4
        & 60.5
        & 63.7
        & 65.0
        & 65.4
        & \textbf{66.5}\\

        & 200$\times$200
        & 58.7
        & 55.7
        & 62.8
        & 55.5
        & 60.5
        & 60.0\\
        
        \noalign{\smallskip}
        \hline
        \noalign{\smallskip}
        
        \multirow{4}{*}{Bright} 
        & 50$\times$50
        & 53.1
        & 53.5
        & 53.1
        & 53.9
        & 57.8
        & 59.5
\\
        
        & 100$\times$100
        & 65.6
        & 66.3
        & 64.1
        & \textbf{67.0}
        & 66.3
        & 65.6
\\
        
        & 150$\times$150
        & 61.3
        & 62.4
        & 64.1
        & 63.1
        & 65.9
        & 63.4
\\

        & 200$\times$200
        & 59.9
        & 59.2
        & 62.4
        & 59.5
        & 58.5
        & 59.2
\\

        \noalign{\smallskip}
        \hline
        \noalign{\smallskip}
        
        \multirow{4}{*}{Dark} 
        & 50$\times$50
        & 53.8
        & 50.5
        & 57.6
        & 54.3
        & 51.6
        & 54.3

\\
        
        & 100$\times$100
        & 61.4
        & \textbf{64.1}
        & 58.6
        & 59.2
        & 61.9
        & 63.5

\\
        
        & 150$\times$150
        & 56.5
        & 61.9
        & 61.4
        & 60.8
        & 61.4
        & 63.5

\\

        & 200$\times$200
        & 54.8
        & 54.8
        & 57.0
        & 57.0
        & 53.8
        & 53.8

\\
        
        \noalign{\smallskip}
        \hline
        
        \end{tabular}
\end{table}


\begin{table}[!tb] 
    \caption{Classification results for 1:2 data subset when varying the epoch value using modified AlexNet}
      \label{table:cnn_12}
      \centering
        \begin{tabular}{cccccccc}
        \noalign{\smallskip}
        \cline{3-8} 
        \noalign{\smallskip}
        
        & & \multicolumn{6}{c}{Epoch}\\
        
        \noalign{\smallskip}
        \cline{2-8} 
        \noalign{\smallskip}
         & Resolution 
         & 100 
         & 120
         & 140
         & 160
         & 180
         & 200	\\
        \noalign{\smallskip}
        \hline
        \noalign{\smallskip}
        
        \multirow{4}{*}{Bright + Dark} 
        & 50$\times$50
        & 63.8
        & 62.2
        & 58.7
        & 62.9
        & 57.7
        & 60.6
\\
        
        & 100$\times$100
        & 62.5
        & 63.3
        & 66.3
        & 66.6
        & 66.3
        & 63.6
\\
        
        & 150$\times$150
        & 63.9
        & 66.8
        & 67.3
        & 65.9
        & \textbf{67.6}
        & 67.5
\\

        & 200$\times$200
        & 62.2
        & 61.2
        & 60.9
        & 62.2
        & 61.3
        & 61.9
\\
        
        \noalign{\smallskip}
        \hline
        \noalign{\smallskip}
        
        \multirow{4}{*}{Bright} 
        & 50$\times$50
        & 50.0
        & 65.2
        & 64.3
        & 62.6
        & 62.4
        & 58.6
\\
        
        & 100$\times$100
        & 64.0
        & 69.7
        & 69.0
        & 66.9
        & 68.5
        & 67.3
\\
        
        & 150$\times$150
        & 65.9
        & 61.2
        & 63.3
        & 65.0
        & 65.0
        & 62.6
\\

        & 200$\times$200
        & \textbf{69.9}
        & 66.4
        & 60.9
        & 63.3
        & 64.7
        & 66.6
\\

        \noalign{\smallskip}
        \hline
        \noalign{\smallskip}
        
        \multirow{4}{*}{Dark} 
        & 50$\times$50
        & 65.9
        & 64.1
        & 64.4
        & 63.7
        & 62.6
        & 63.4
\\
        
        & 100$\times$100
        & 64.8
        & 67.0
        & 67.7
        & 68.4
        & \textbf{69.9}
        & 67.7
\\
        
        & 150$\times$150
        & 62.3
        & 60.8
        & 61.9
        & 63.7
        & 63.4
        & 60.1
\\

        & 200$\times$200
        & 59.0
        & 61.5
        & 63.0
        & 61.5
        & 59.4
        & 63.0
\\
        
        \noalign{\smallskip}
        \hline
        
        \end{tabular}
\end{table}


\begin{table}[!tb] 
    \caption{Classification results for 1:3 data subset when varying the epoch value using modified AlexNet}
      \label{table:cnn_13}
      \centering
        \begin{tabular}{cccccccc}
        \noalign{\smallskip}
        \cline{3-8} 
        \noalign{\smallskip}
        
        & & \multicolumn{6}{c}{Epoch}\\
        
        \noalign{\smallskip}
        \cline{2-8} 
        \noalign{\smallskip}
         & Resolution 
         & 100 
         & 120
         & 140
         & 160
         & 180
         & 200	\\
        \noalign{\smallskip}
        \hline
        \noalign{\smallskip}
        
        \multirow{4}{*}{Bright + Dark} 
        & 50$\times$50
        & 72.6
        & 68.6
        & 68.7
        & 65.3
        & 66.6
        & 68.0
\\
        
        & 100$\times$100
        & 67.7
        & 71.6
        & 72.3
        & 72.2
        & 71.8
        & 73.4
\\
        
        & 150$\times$150
        & \textbf{74.0}
        & 72.0
        & \textbf{74.0}
        & 71.6
        & 71.1
        & 73.1
\\

        & 200$\times$200
        & 68.4
        & 70.2
        & 70.1
        & 70.1
        & 71.2
        & 70.6
\\
        
        \noalign{\smallskip}
        \hline
        \noalign{\smallskip}
        
        \multirow{4}{*}{Bright} 
        & 50$\times$50
        & 75.0
        & 74.1
        & 71.4
        & 70.0
        & 64.3
        & 63.4
\\
        
        & 100$\times$100
        & 74.8
        & 74.2
        & 72.5
        & 67.0
        & 68.2
        & 65.6
\\
        
        & 150$\times$150
        & 71.4
        & 72.8
        & 73.7
        & 74.4
        & \textbf{76.2}
        & 75.1
\\

        & 200$\times$200
        & 73.4
        & 73.5
        & 73.5
        & 74.6
        & 73.4
        & 75.1
\\

        \noalign{\smallskip}
        \hline
        \noalign{\smallskip}
        
        \multirow{4}{*}{Dark} 
        & 50$\times$50
        & 75.0
        & 72.2
        & 69.5
        & 67.9
        & 67.3
        & 67.6
\\
        
        & 100$\times$100
        & 67.6
        & 69.8
        & 72.5
        & 73.6
        & 72.2
        & 70.9
\\
        
        & 150$\times$150
        & 72.5
        & 66.8
        & 69.8
        & 70.9
        & 70.6
        & 69.5
\\

        & 200$\times$200
        & 73.3
        & 68.2
        & 70.9
        & 72.8
        & 73.3
        & \textbf{74.1}
\\
        
        \noalign{\smallskip}
        \hline
        
        \end{tabular}
\end{table}


\begin{table}[!tb] 
    \caption{Confusion matrix by adopting modified AlexNet for 1:3 ``bright+dark" data subset, when epoch=140 and resolution=150$\times$150}\label{table:confusion_13_brightdark}
      \centering
        \begin{tabular}{lcc}
        \noalign{\smallskip}
        \cline{1-3} 
        \noalign{\smallskip}
         &	Non-defective   & Defective 	\\
        \noalign{\smallskip}
        \hline
        \noalign{\smallskip}
         Non-defective
        & 616	& 83 	\\
         Defective 
        & 159 & 74   	\\
        \hline
        \end{tabular}
\end{table}


\begin{table}[!tb] 
    \caption{Confusion matrix by adopting modified AlexNet for 1:3 ``bright" data subset, when epoch=180 and resolution=150$\times$150}\label{table:confusion_13_bright}
      \centering
        \begin{tabular}{lcc}
        \noalign{\smallskip}
        \cline{1-3} 
        \noalign{\smallskip}
         &	Non-defective   & Defective 	\\
        \noalign{\smallskip}
        \hline
        \noalign{\smallskip}
         Non-defective
        & 381	& 42 	\\
         Defective 
        & 92 & 49   	\\
        \hline
        \end{tabular}
\end{table}


\begin{table}[!tb] 
    \caption{Confusion matrix by adopting modified AlexNet for 1:3 ``dark" data subset, when epoch=200 and resolution=200$\times$200}\label{table:confusion_13_dark}
      \centering
        \begin{tabular}{lcc}
        \noalign{\smallskip}
        \cline{1-3} 
        \noalign{\smallskip}
         &	Non-defective   & Defective 	\\
        \noalign{\smallskip}
        \hline
        \noalign{\smallskip}
         Non-defective
        & 250	& 26 	\\
         Defective 
        & 69 & 23   	\\
        \hline
        \end{tabular}
\end{table}

\section{Conclusion}\label{sec:conclusion}
This paper presents two neural network approaches to distinguish the defective and non-defective leather images:  Artificial Neural Network (ANN) and Convolutional Neural Network (CNN).
For ANN, there are four pre-processing steps involved before extracting the features from the images.
They include RGB to grayscale, re-sizing, edge detection and block partitioning.
All the images are put through these steps to improve their quality and to eliminate noise.
Then, the images are passed to ANN to further select effective features to represent the image.
Experimental results show that the proposed method achieves a promising classification accuracy of 80\%. 

On the other hand, for CNN, only good samples are selected for evaluation. 
To reduce the impact of data class imbalance issue, the dataset is reconstructed to form 1:1, 1:2, 1:3 data distributions for defective:non-defective images.
As a result, the highest classification accuracy obtained is 76\% when the images is resized to more than half of the original size.
The features are extracted using modified AlexNet with relatively few training epoch to fine-tune the weights and biases in the architecture.

As future work, more defective leather samples can be added to the testing dataset in the experiment to avoid the feature extractor learns the features of one particular class.
Moreover, the number of hidden neurons and the number of feature vectors in ANN can be increased to obtain more accurate results.
Besides, popular pre-trained CNN models such as GoogLeNet, SqueezeNet, VGG-16, ResNet-101 can be employed to extract the important features of the image and hence generating higher classification results.

\section*{Acknowledgments}
This work was funded by Ministry of Science and Technology (MOST) (Grant Number: MOST 107-2218-E-035-016-), National Natural Science Foundation of China (No. 61772023) and Natural Science Foundation of Fujian Province (No. 2016J01320).

\bibliographystyle{natlib}
\bibliography{ref}

\end{document}